\pdfoutput=1

\documentclass[11pt]{article}
\usepackage{multirow}
\usepackage{svg}

\usepackage[final]{acl}

\usepackage{times}
\usepackage{latexsym}

\usepackage[T1]{fontenc}

\usepackage[utf8]{inputenc}

\usepackage{microtype}
\usepackage{booktabs}
\usepackage{inconsolata}

\usepackage{graphicx}
\usepackage{enumitem}
\usepackage{xcolor}
\definecolor{mybgcolor}{HTML}{ea998f} 
\definecolor{mybgcolor2}{HTML}{c4d1f7}
\definecolor{mybgcolor3}{HTML}{ffecc1}
\definecolor{mybgcolor4}{HTML}{aec5a8}

\definecolor{color1a}{RGB}{93, 116, 162}
\definecolor{color1b}{RGB}{196, 216, 242}
\definecolor{color2a}{RGB}{142, 45, 48}
\definecolor{color2b}{RGB}{242, 232, 227}

\definecolor{myblue}{RGB}{20, 76, 115}    
\definecolor{myorange}{RGB}{163, 82, 9}       
\definecolor{mygreen}{RGB}{28, 102, 28}     
\definecolor{myred}{RGB}{137, 25, 26}    

\title{Distilling Empathy from Large Language Models}

\author{
  Henry J.~Xie\textsuperscript{1}, 
  Jinghan Zhang\textsuperscript{2}, 
  Xinhao Zhang\textsuperscript{2}, 
  Kunpeng Liu\textsuperscript{2} \\
  \textsuperscript{1}Westview High School, Portland, OR 97229, USA \\
  \texttt{henryjxie@gmail.com} \\
  \textsuperscript{2}Portland State University, Portland, OR 97201, USA \\
  \texttt{\{jinghanz, xinhaoz, kunpeng\}@pdx.edu}
}

\begin{document}
\maketitle

\begin{abstract}
The distillation of knowledge from Large Language Models (LLMs) into Smaller Language Models (SLMs), preserving the capabilities and performance of LLMs while reducing model size, has played a key role in the proliferation of LLMs.
Because SLMs are considerably smaller than LLMs, they are often utilized in domains where human interaction is frequent but resources are highly constrained, e.g., smart phones. Therefore, it is crucial to ensure that empathy, a fundamental aspect of positive human interactions, already instilled into LLMs, is retained by SLMs after distillation.
In this paper, we develop a comprehensive approach for effective empathy distillation from LLMs into SLMs. Our approach features a two-step fine-tuning process that fully leverages datasets of empathetic dialogue responses distilled from LLMs. We explore several distillation methods beyond basic direct prompting and propose four unique sets of prompts for targeted empathy improvement to significantly enhance the empathy distillation process. Our evaluations demonstrate that SLMs fine-tuned through the two-step fine-tuning process with distillation datasets enhanced by the targeted empathy improvement prompts significantly outperform the base SLM at generating empathetic responses with a win rate of 90+\%. Our targeted empathy improvement prompts substantially outperform the basic direct prompting with a 10+\% improvement in win rate.\footnote{Code Repository: https://github.com/henryjxie/Distilling-Empathy-from-Large-Language-Models}
\end{abstract}

\section{Introduction}

\begin{figure}[htb]
    \vspace{-0.05in}
    \centering
    \includegraphics[width=1\linewidth]{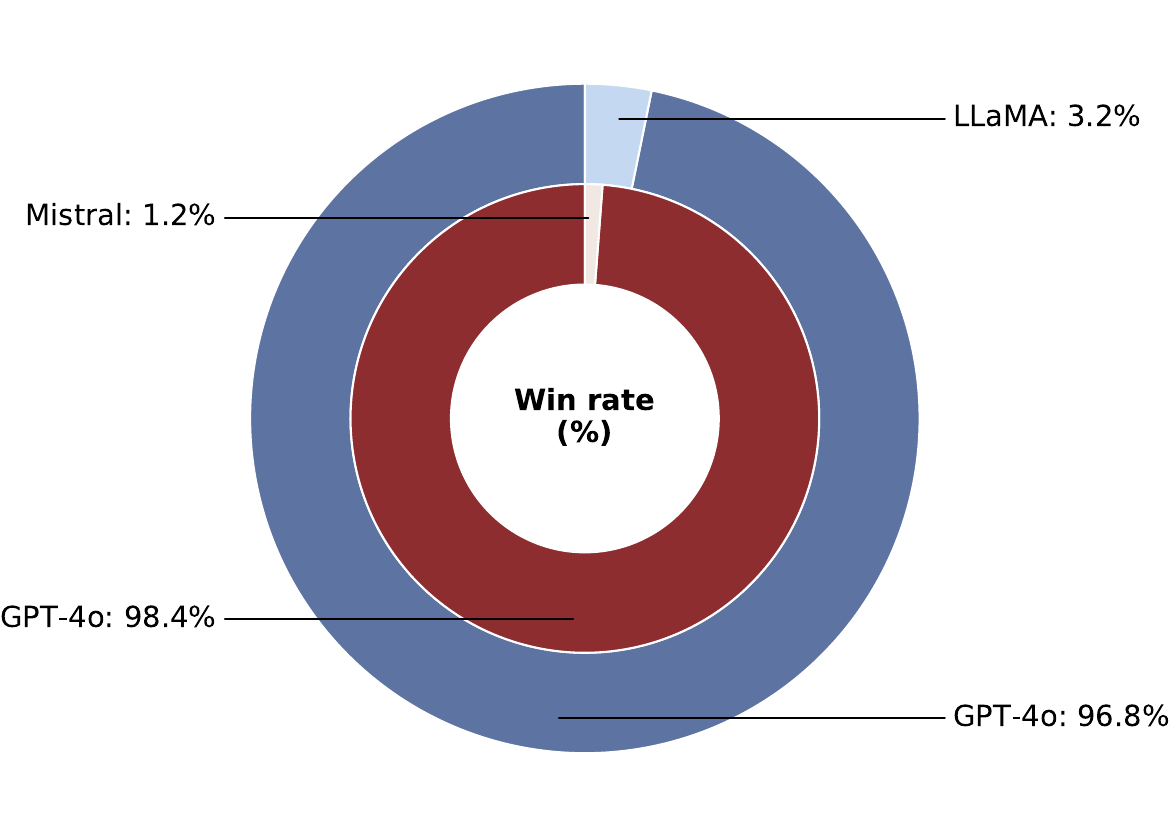}
    \vspace{-0.15in}
    \caption{GPT-4o vs. Base LLaMA-3.1-8B \& Mistral-7B-v0.3 in empathetic responses as judged by Gemini}
    \label{fig:win-rate-comp}
    \vspace{-0.15in}
\end{figure}

An emerging trend in the development and application of Large Language Models (LLMs) is the proliferation of Smaller Language Models (SLMs)~\cite{DeepSeekAI2025DeepSeekR1IR}. SLMs are essential to the widespread adoption of LLMs as their significantly smaller sizes allow them to be employed in many  application settings that are highly human-interactive but resource-constrained, such as smart phones and intelligent home devices~\cite{cui2023rise, qualcomm2023ondevice, Wang_2025}.

A popular method for developing the capabilities of SLMs is knowledge distillation from LLMs, ensuring that the smaller models perform nearly at the same level as the larger models while being much more efficient in size and resource needs~\cite{Xu2024ASO}. SLMs have been shown to perform very well in specific tasks for which they are distilled~\cite{Sreenivas2024LLMPA}. Due to their close interactions with humans, it is highly desirable for SLMs to have proficient empathetic abilities, which are essential for successful positive interactions with humans. In recent years, LLMs have been shown to possess an impressive understanding of empathy, similar to or surpassing humans' comprehension of empathy~\cite{Welivita2024AreLL}. However, as shown in Figure~\ref{fig:win-rate-comp}, base SLMs such as LLaMA-3.1-8B~\cite{meta2024LLaMA3} and Mistral-7B-v0.3~\cite{mistral2023mistral7b} perform considerably worse in empathy than a state-of-the-art LLM such as GPT-4o~\cite{openai2024gpt4o}. GPT-4o outperforms these base SLMs with win rates above 96\% in generating empathetic responses as judged by Gemini-2.0-Flash~\cite{google2024gemini2}. Therefore, it is strongly desired that the level of empathy possessed by LLMs can be preserved during the distillation.

In this paper, we develop a comprehensive approach for systematically distilling empathy from LLMs into SLMs with three key components: 
\begin{itemize}
    \item {\bf Two-step fine-tuning.} We develop a two-step fine-tuning process: first supervised fine-tuning (SFT) and then reinforcement learning with human feedback (RLHF) by direct preference optimization (DPO)~\cite{rafailov2023dpo}. Using empathetic responses distilled from four state-of-the-art LLMs (GPT-4, LLaMA, Gemini, and Mixtral)~\cite{Welivita2024AreLL}, this process utilizes both high and low empathy responses in fine-tuning: high empathy responses for SFT and (low, high) empathy response pairs for RLHF DPO, leveraging human empathy scores provided in the dataset. The results show that such fine-tuning significantly improves the empathy of SLMs. 
    \item {\bf Three empathy distillation methods.} Besides (1) the {\em basic direct distillation method}---simply asking LLMs to generate empathetic responses given dialogue contexts---we explore two additional distillation methods: (2) {\em targeted empathy improvement over human responses:} given initial human responses, LLMs are instructed to improve the empathy of the responses with different prompting strategies and the improved responses and initial human responses are utilized to construct the SFT and RLHF datasets for the two-step fine-tuning process; (3) {\em targeted empathy improvement over LLM initial responses:} initial responses are generated by LLMs instead of humans, and LLMs are then instructed to improve over the initial responses. Method 2 anchors empathy improvement in human responses while Method 3 bootstraps the distillation datasets without human involvement.  
    \item {\bf Four prompting strategies for empathy distillation.} We propose and implement four different sets of prompts for targeted empathy improvement to significantly enhance the empathy distillation process. These strategies prompt LLMs to improve initial responses along the three dimensions of empathy (cognitive, affective, and compassionate dimensions), exploring different ways to emphasize or combine these empathy dimensions. The resulting improved responses can then be utilized in our two-step fine-tuning process for SLMs. The initial responses and the improved ones are split into the SFT and RLHF datasets based on human empathy scores (if available) or prior successful partitioning ratios.
\end{itemize}
Our evaluations show that SLMs fine-tuned by our two-step process, with distillation datasets enhanced by targeted empathy improvement prompts, significantly outperform the base SLMs at generating empathetic responses with a win rate of 90+\% over the base. Our targeted empathy improvement prompts substantially outperform the basic direct prompting with a 10+\% improvement in win rate.

\begin{figure*}
\begin{center}
    \fbox{\rule[-.15cm]{0cm}{0cm}
    \begin{minipage}[t]{6in}
       {\small
        \noindent{\bf Sample Dialogue and Responses:}
        \\
        {\bf Context:} I was just walking out of Taco bell. The bottom fell out of my bag and my wonderful taco bell covered the ground.
        \\
        {\bf Speaker Utterance:} I can't believe the bottom of the bag ripped and my wonderful taco's covered the ground. I'm sure my face was bright red with anger.
        \\

        {\bf Human Response:} NOT THE TACOS!!!! Seriously, I can understand how frustrating that is. Especially when you have the taste for something.....then gone!
        \\
        {\bf Human Response Empathy Score:} 1
        \\

        {\bf GPT-4 Response:} I'm really sorry to hear about your tacos. It's so frustrating when something you're looking forward to doesn't go as planned. I can understand why you'd be upset.
        \\
        {\bf GPT-4 Response Empathy Score:} 3
        \\

        {\bf LLaMA Response:} Oh no, that's frustrating! I can understand how you must have felt, especially when you were looking forward to enjoying your tacos. It's embarrassing when things like that happen unexpectedly.
        \\
        {\bf LLaMA Response Empathy Score:} 2
        \\

        {\bf Gemini Response:} I understand your frustration and disappointment. It can be incredibly upsetting when something like that happens, especially when you were looking forward to enjoying your tacos. It's natural to feel angry and embarrassed in such a situation.
        \\
        {\bf Gemini Response Score:} 3
        \\

        {\bf Mixtral Response:} I'm so sorry to hear that your tacos spilled! That can be so frustrating, I've been there too. It's like the worst feeling when you're looking forward to something delicious and it doesn't go as planned. Is there anything I can do to help you get a new taco?
        \\
        {\bf Mixtral Response Empathy Score:} 3
}
    \end{minipage}
}
\end{center}
\caption{Sample dialogue with responses from humans and LLMs, and human empathy scores for each response}
\label{fig:sample-dial}
\end{figure*}

\section{Background}

\subsection{LLM Distillation}

Model distillation has played a key role in the proliferation of LLMs and SLMs~\cite{Xu2024ASO}, transferring the advanced capabilities of a larger model to a smaller model. There has been extensive previous work on LLM distillation~\cite{Sreenivas2024LLMPA}. There are two major complementary parts of distillation: (1) extracting high quality data from an LLM, which reflects the advanced capability of the LLM that is desired to be transferred, and (2) effectively instilling the advanced capability into the target SLM by constructively utilizing the extracted data through appropriate model fine-tuning techniques. A well-known success story of this type of model distillation is DeepSeek's work on distilling the advanced reasoning capability of its DeepSeek-R1 model into several state-of-the-art SLMs, such as LLaMA~\cite{DeepSeekAI2025DeepSeekR1IR}. Some distilled SLMs even outperform DeepSeek-R1 in certain reasoning tasks while being an order of magnitude smaller in size than DeepSeek-R1.

\subsection{Related Work on Empathetic Responses}

There has been much research on empathy in dialogue systems
~\cite{Ma2020, Raamkumar2022}. 
One such work is on the instillation of empathetic intents into language models to guide the generation of empathetic responses~\cite{welivita-pu-2020-taxonomy}. 
With the advancement in LLMs, the attention has been turned to the empathetic capabilities of LLMs~\cite{Welivita2024AreLL}. There has also been research on using LLMs to measure empathy of dialogue responses~\cite{Xie2024ScoringWL}. Another closely related work is on emotion guided paraphrasing~\cite{DBLP:conf/wassa/XieA23}, which can paraphrase an initial response along certain emotion gradients to potentially show better empathy.

\section{Dataset Statistics and Analysis}

In this study, we use the LLMs-vs-Humans dataset created by~\cite{Welivita2024AreLL}. It is sourced from the Empathetic Dialogues dataset~\cite{Rashkin2018TowardsEO} and contains 2000 unique dialogue contexts (situation and speaker utterance). As shown in Figure~\ref{fig:sample-dial}, for each context, there are a human response to the speaker utterance and the responses from four LLMs: GPT-4~\cite{openai2023gpt4}, LLaMA-2-70B-Chat~\cite{touvron2023LLaMA}, Gemini-1.0-Pro~\cite{google2023gemini}, and Mixtral-8x7B-Instruct~\cite{mistralai2024mixtral}. For each response, a human-annotated empathy score is given with the scale: \textit{1--bad empathetic response, 2--okay, 3--good}. 

%
%

\begin{figure}[p]
    \centering
    \includegraphics[width=1\linewidth]{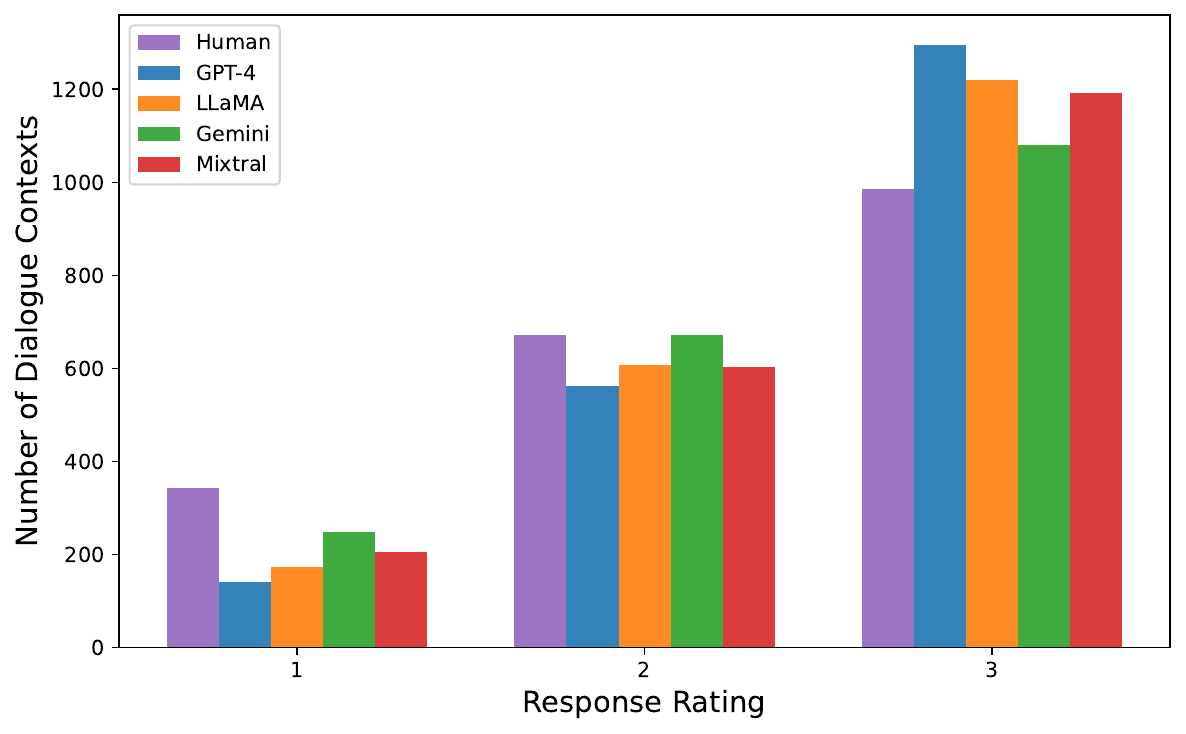}
    \caption{Empathy score distribution}
    \label{fig:emp-score-dist}

    \vspace{0.15in}
    \centering
    \includegraphics[width=1\linewidth]{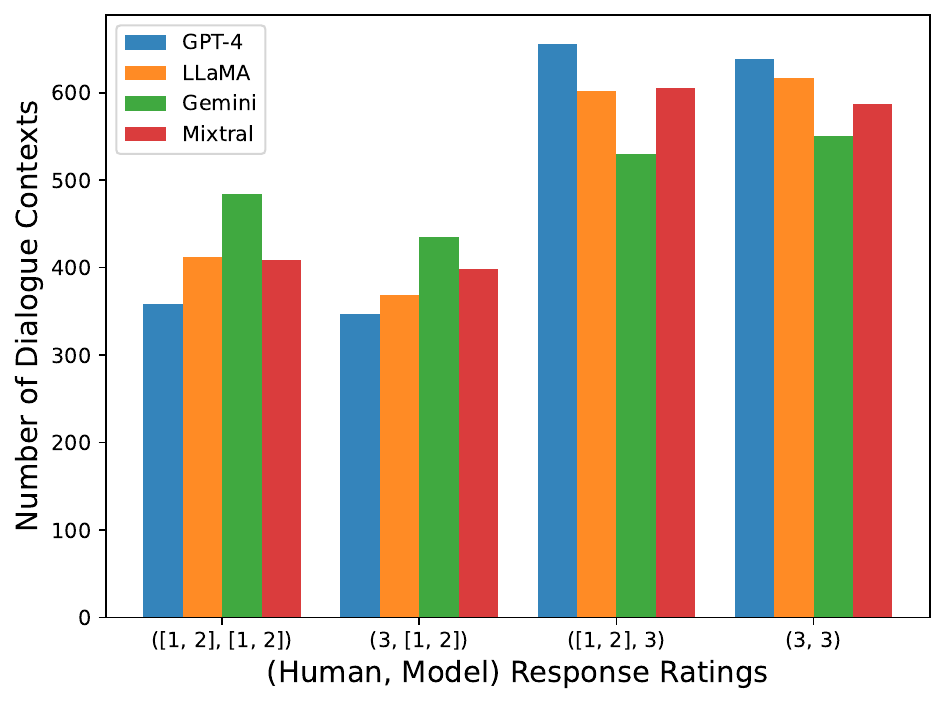}
    \caption{Distribution of (human, model) response empathy rating pairs} 
    \label{fig:empathy-score-correlation}
    \vspace{-0.15in}
\end{figure}

Figure~\ref{fig:emp-score-dist} shows the empathy score distribution of the responses from each responder in the dataset. The majority of responses from every responder received an empathy score of 3, indicating that as the dataset was created, the prompts given to humans and LLMs were effective in extracting empathetic responses. LLMs were, in general, more empathetic than humans when responding to the given dialogues as noted by \cite{Welivita2024AreLL}.  
Figure~\ref{fig:empathy-score-correlation} illustrates the number of dialogue contexts for each (human, model) response empathy rating pair. On one hand, there are many pairs of human and LLM responses that both received empathy scores of 3, i.e., good empathetic response, suitable for SFT. On the other hand, there are also many pairs with differing empathy scores. Particularly interesting are those pairs where one response received a score of 3 while the other did not, which have potential to serve as contrastive pairs for RLHF.

\section{Two-Step Fine-Tuning for Distillation}

\begin{figure}[htb]
    \centering
    \includegraphics[width=\linewidth]{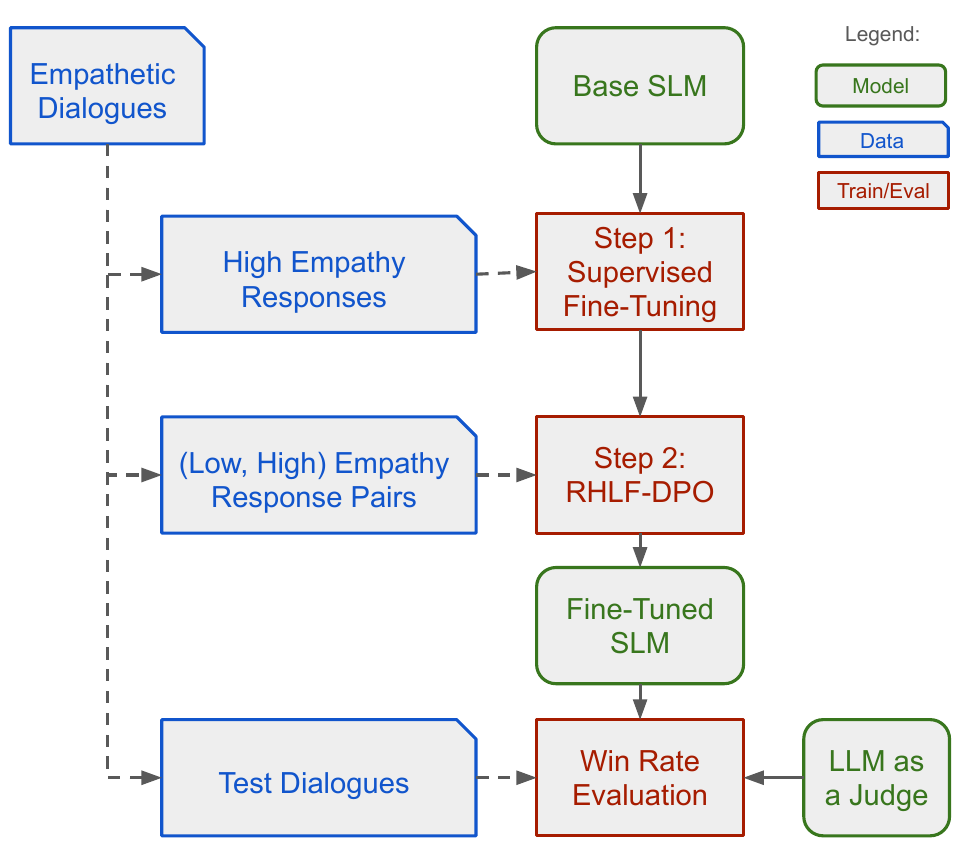}
    \caption{Two-step fine-tuning: SFT then RLHF DPO}
    \label{fig:sft-dpo}
    \vspace{-0.15in}
\end{figure}

Figure~\ref{fig:sft-dpo} illustrates the two-step fine-tuning process that is central to our approach for distilling empathy from LLMs into SLMs. Given the dialogue dataset with responses generated by a human or an LLM, three separate datasets are created for SFT, RLHF, and evaluation. The first fine-tuning step is SFT on the base SLM and the second step is RLHF with DPO on the SLM after SFT. Finally, the fine-tuned SLM is evaluated head-on against the base SLM to measure its improvement using the win rate metric. 

\subsection{SFT and RLHF Dataset Preparation}
\label{SubSec:DatasetPrep}
To explain the procedure of preparing the SFT and RLHF datasets, we utilize the LLMs-vs-Humans dataset by~\cite{Welivita2024AreLL} discussed above as the source for sample empathetic responses. For the SFT dataset, we select all dialogues whose human response and LLM response both received an empathy score of 3. Each human or LLM response with the dialogue context forms an example in the SFT dataset. This dataset includes high empathy response examples from both humans and LLMs. For the RLHF dataset, we select all dialogues whose human response received an empathy score of 1 or 2  while the LLM response received an empathy score of 3. Each pair of human and LLM responses with the dialogue context form a chosen-rejected entry in the dataset. Therefore, the RLHF dataset includes contrastive pairs of (low, high) empathy responses, where the chosen empathetic response is the LLM response which improves over the rejected human response. The remaining dialogues are included in the test dataset for evaluation.

\subsection{Two-Step Fine-Tuning} 
\label{SubSec:SFT-DPO}
Our two-step fine-tuning process operates as follows. We first fine-tune the SLM through SFT with high empathy responses, then fine-tune it further with RLHF DPO. The RLHF dataset contains (low, high) empathy response pairs, where the SLM can learn the chosen high empathy response and recognize the rejected low empathy response. With this process, we can leverage the advantages of both methods. SFT aligns the model with high empathy responses, providing a baseline for reliable and reasonable outputs. RLHF DPO then helps introduce the model to nuance in empathetic responses. With this combination, our fine-tuned model can significantly outperform the base SLM and improve over the SLM fine-tuned only through SFT.

For both steps, we utilize LLaMA Factory, a platform for model fine-tuning~\cite{Zheng2024LLaMAFactoryUE}. We employ the following hyper-parameters for both SFT and RLHF DPO: {\em Fine-Tuning Method = lora, Lora Rank = 8, Learning Rate = 5-e5, Epochs = 3.0, Compute Type = bf16, and Batch Size = 2}. For DPO specifically, we employ the hyper parameters: {\em Beta Value = 0.1 and Loss Type = sigmoid}. All fine-tuning is done using an NVIDIA 4090 GPU.  



\subsection{Distilled Model Evaluation: Win Rate}
\label{SubSec:Win-Rate}
To evaluate the performance of our fine-tuned SLMs, we utilize the evaluation metric of win rate. The win rate metric reflects the percentage of times a fine-tuned SLM's response is preferred over the base SLM's response in terms of empathy, decided by an outside judge. In our study, we employ both GPT-4o and Gemini-2.0-Flash as win rate judges, invoking them through APIs, as they represent the state-of-the-art in LLMs. We calculate the win rate of the SLM fine-tuned through SFT then RLHF DPO over the base SLM and compare it with the win rate of the SLM fine-tuned only through SFT over the base SLM. The use of the win rate metric allows us to easily recognize which model has a larger improvement in empathy over the base SLM.

\section{Empathy Distillation Methods}
We explore three methods for distilling empathy from LLMs: (1) direct empathy distillation, (2) targeted empathy improvement over human responses, and (3) targeted empathy improvement over LLM initial responses. Method 1 differs from Method 2 in that Method 1 prompts the LLM to directly generate a response while Method 2 asks the LLM to improve over a human response. Method 3 combines aspects from Methods 1 and 2 to prompt the LLM to improve over an LLM initial response.

\subsection{Method 1: Direct Empathy Distillation}

In direct empathy distillation, LLMs are queried with a simple and straightforward prompt that asks them to generate an empathetic response for a given dialogue context (situation and speaker utterance). This exercise has been done by \cite{Welivita2024AreLL} when creating their LLMs-vs-Humans dataset. 
Four different LLMs are queried with the basic direct prompt to generate responses for the 2000 dialogue contexts. This essentially creates a dataset distilling empathy from these LLMs. We use this dataset to investigate the effectiveness of direct empathy distillation by creating SFT and RLHF datasets for each of the four state-of-the-art LLMs as discussed in Section~\ref{SubSec:DatasetPrep}. We combine the SFT and RLHF datasets of all four LLMs into a combined SFT dataset and a combined RLHF dataset for measuring the maximal effect that can be achieved with this pre-existing dataset. To distill the empathy of the four state-of-the-art LLMs into SLMs, we employ the two-step fine-tuning process presented in Section~\ref{SubSec:SFT-DPO}. We fine-tune an SLM five times, one for each LLM and the combined dataset of all four LLMs. Subsequently, we evaluate the fine-tuned SLMs in terms of win rate over the base SLM using GPT-4o as the judge, as discussed in Section~\ref{SubSec:Win-Rate} (same setup used for Methods 2 and 3).

\begin{figure}[htb]
    \centering
    \includegraphics[width=1\linewidth]{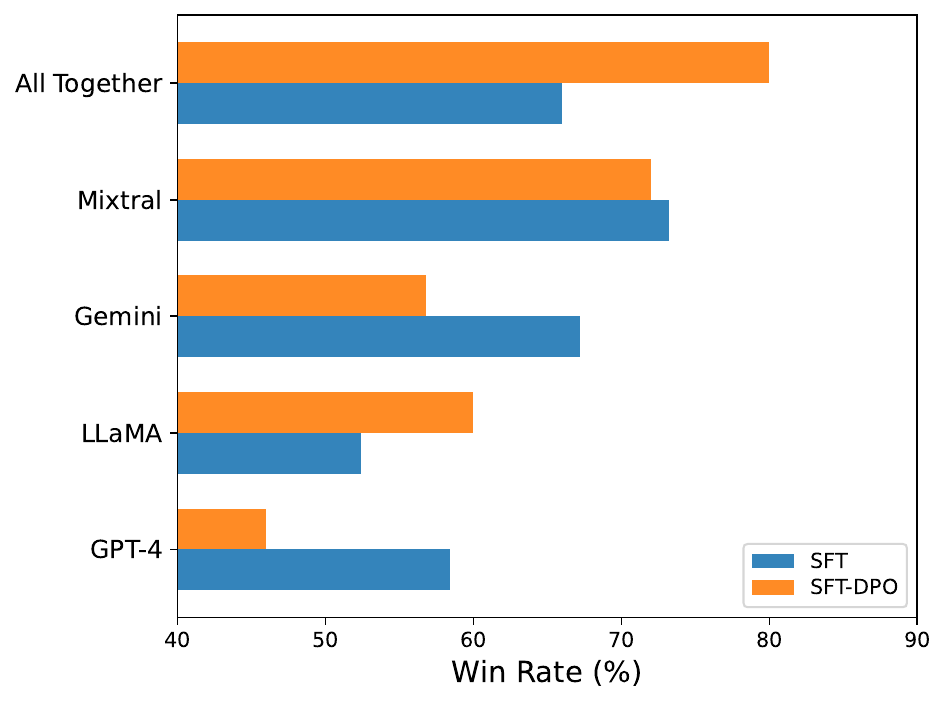}
    \caption{Performance of direct empathy distillation}
    \label{fig:direction-distillation}
    \vspace{-0.2in}
\end{figure}

Figure~\ref{fig:direction-distillation} illustrates the performance of direct distillation from the four LLMs, GPT-4, LLaMA, Gemini, and Mixtral, into a SLM, namely LLaMA-3.1-8B. It can be observed that SFT-only leads to fine-tuned models that achieve consistently better empathy performance over the base SLM. However, SFT then RLHF DPO has uneven performance with datasets from individual LLMs. The dataset that combines all four LLM datasets achieves the best performance over the base SLM through SFT then RLHF DPO. On the other hand, SFT with the combined dataset even under-performs the datasets of some individual LLMs. In summary, direct empathy distillation can achieve major improvement over the base SLM; however, its effectiveness is uneven and including more examples into the SFT dataset does not necessarily guarantee improvement in performance of the fine-tuned SLMs.

\subsection{Method 2: Targeted Empathy Improvement over Human Responses}
Direct empathy distillation shows potential in effectively distilling empathy from LLMs into SLMs. However, it utilizes a generic prompting style for distillation and is not tailored specifically for distilling empathy. In this section, we develop an empathy specific distillation method. The core of this method is a set of prompts for targeted empathy improvement over a human response. To serve as the baseline, we first develop a naive prompt for empathy improvement, which is derived from the direct empathy distillation prompt used by \cite{Welivita2024AreLL} when creating their dataset. Instead of directly asking for an empathetic response, this naive prompt asks the LLM to improve a given response.
\begin{center}
    \fbox{\rule[-.15cm]{0cm}{0cm}
    \begin{minipage}[t]{2.8in}
       {\small
        \noindent{\bf Naive Prompt for Empathy Improvement}
        \\
        Below is a response to a given speaker utterance in a given context. \textcolor{myblue}{Generate a new improved empathetic response,} using on average 28 words and a maximum of 97 words, that is of \textcolor{myblue}{higher empathetic quality} and also \textcolor{myblue}{retains the original meaning, intention, and emotion} of the original response.
        }
    \end{minipage}
}
\end{center}

We develop four different prompts for targeted empathy improvement along the three dimensions of empathy: cognitive, affective, and compassionate empathy as established by psychologists~\cite{davis1983}. 
The targeted empathy improvement prompts that we design follow the structure shown below. 
\begin{center}
    \fbox{\rule[-.15cm]{0cm}{0cm}
    \begin{minipage}[t]{2.8in}
       {\small
        \noindent{\bf Prompt Name}
        \\
        \{Naive prompt\}
        \\
        \{Empathy improvement strategy\}
        \\
        \{Definitions of empathy dimensions used by the strategy\}
        }
    \end{minipage}
}
\end{center}
It first includes the naive prompt, then defines the strategy for empathy improvement, and finally provides the definitions of different dimensions of empathy as needed. This structure makes prompt design uniform, facilitating our evaluation. The naive prompt also follows this structure, albeit with no strategy or empathy dimension definitions.   

The first prompt is a set of prompts that instruct the LLM to improve the response along one specific dimension of empathy. There is one prompt for improving the cognitive dimension of empathy only, one for affective, and one for compassionate. 
\begin{center}
    \fbox{\rule[-.15cm]{0cm}{0cm}
    \begin{minipage}[t]{2.8in}
        {\small
        \noindent{\bf Prompt 1.1: Improve along the Cognitive Dimension}\\[1ex]
        \{Naive Prompt\}
        \\
        \{Strategy\} Your higher quality response should be \textcolor{myblue}{improved specifically along the cognitive dimension} of empathy. 
        \\
        \{Definition of cognitive dimension of empathy\} 
        Cognitive empathy is \textcolor{mygreen}{the ability to understand another person's thoughts, beliefs, and intentions. It is being able to see the world through their eyes and understand their point of view.}
        \\[1ex]
        \noindent{\bf Prompt 1.2: Improve along the Affective Dimension}\\[1ex]
        \{Naive Prompt\}
        \\
        \{Strategy\} Your higher quality response should be \textcolor{myblue}{improved specifically along the affective (emotional) dimension} of empathy. 
        \\
        \{Definition of affective dimension of empathy\} 
        Affective empathy is \textcolor{mygreen}{the ability to experience the emotions of another person. It is feeling what they are feeling, both positive and negative.}
        \\[1ex]
        \noindent{\bf Prompt 1.3: Improve along the Compassionate Dimension}\\[1ex]
        \{Naive Prompt\}
        \\
        \{Strategy\} Your higher quality response should be \textcolor{myblue}{improved specifically along the compassionate dimension} of empathy. 
        \\
        \{Definition of compassionate dimension of empathy\} 
        Compassionate empathy is \textcolor{mygreen}{the ability to not only understand and share another person's feelings, but also to be moved to help if needed. It involves a deeper level of emotional engagement than cognitive empathy, prompting action to alleviate other's distress or suffering.}
        }
    \end{minipage}
}
\end{center}
With each prompt, we instruct the LLM to generate the improved response for each dialogue, which form the dataset for evaluating this prompt. 
As in Method 1, with the improved responses, we fine-tune the SLM with the two-step process and evaluate the fine-tuned SLMs with the win rate metric.

The second prompt instructs the LLM to improve a response in all three dimensions of empathy. This prompt provides the LLM with information on what each dimension of empathy represents, and the LLM can then improve the response with how it understands all three dimensions.
\begin{center}
    \fbox{\rule[-.15cm]{0cm}{0cm}
    \begin{minipage}[t]{2.8in}
       {\small
        \noindent{\bf Prompt 2: Improve All Three Dimensions of Empathy}
        \\
        \{Naive Prompt\}
        \\
        \{Strategy\} Your higher quality response should be \textcolor{myblue}{improved along the three dimensions of empathy:} cognitive, affective (emotional), and compassionate empathy.
        \\
        \{Definitions of the cognitive, affective, and compassionate dimensions of empathy\}
        }
    \end{minipage}
}
\end{center}

The third prompt contains the sequential application of three different prompts. The LLM is instructed to first improve a response on the cognitive dimension of empathy using Prompt 1.1, then in a second call to further improve on the affective dimension using Prompt 1.2, and finally in a third call to further improve on the compassionate dimension using Prompt 1.3. This three-step prompting possibly allows for a more systematic improvement of the response along three dimensions of empathy.
\begin{center}
    \fbox{\rule[-.15cm]{0cm}{0cm}
    \begin{minipage}[t]{2.8in}
       {\small
        \noindent{\bf Prompt 3: Improve Three Dimensions Sequentially}\\[1ex]
        Apply \{Prompt 1.1\} to the input response 
        \\
        Apply \{Prompt 1.2\} to the output response of 1.1 
        \\
        Apply \{Prompt 1.3\} to the output response of 1.2
        }
    \end{minipage}
}
\end{center}

The fourth prompt instructs the LLM to first identify the dimension of empathy that the given response lacks the most, and then improve on that dimension. This prompt directs the LLM to take a nuanced view of the response in terms of empathy. 
\begin{center}
    \fbox{\rule[-.15cm]{0cm}{0cm}
    \begin{minipage}[t]{2.8in}
       {\small
        \noindent{\bf Prompt 4: Identify the Lacking Dimension}
        \\
        \{Naive Prompt\}
        \\
        \{Strategy\} In the process of generating a higher quality empathetic response, you should \textcolor{myblue}{identify the dimension of empathy} (cognitive, affective, and compassionate dimensions) \textcolor{myblue}{that the original response lacks most of,} and \textcolor{myblue}{specifically improve along the lines of the dimension you identified}.
        \\
        \{Definitions of the cognitive, affective, and compassionate dimensions of empathy\}
        }
    \end{minipage}
}
\end{center}

\begin{figure}[htb]
    \centering
    \includegraphics[width=1\linewidth]{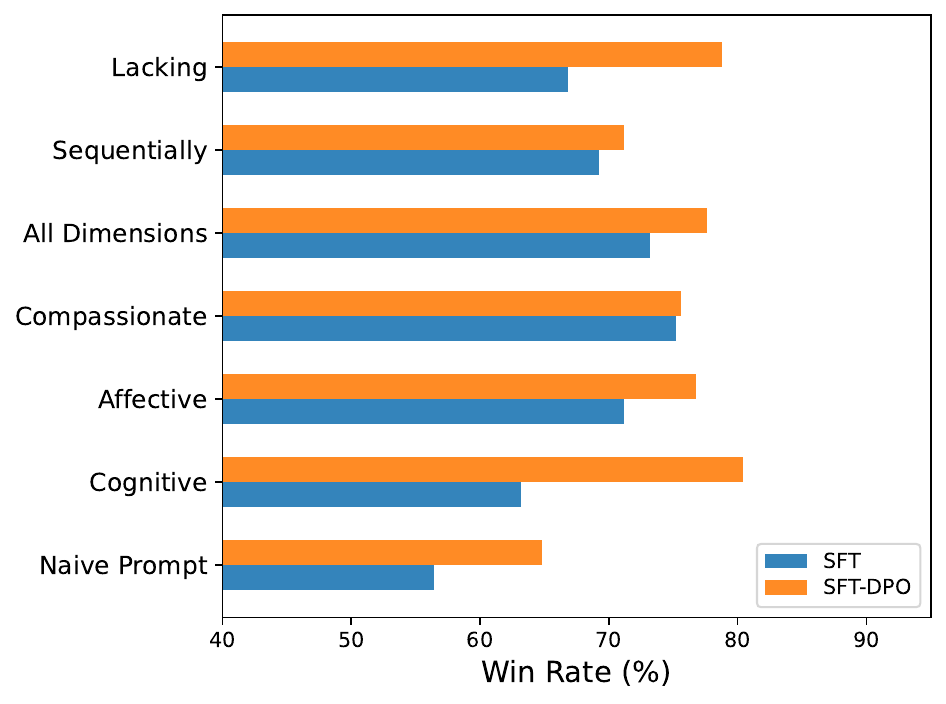}
    \caption{Performance of targeted empathy improvement over human responses}
    \label{fig:targeted-empathy-improvement}
    \vspace{-0.2in}
\end{figure}
Figure~\ref{fig:targeted-empathy-improvement} illustrates the performance of targeted empathy improvement over human responses, with GPT-4o as the LLM and LLaMA-3.1-8B as the SLM. The fine-tuned SLMs for all prompts substantially improve over the base SLM. Three prompts closely match the top-line (all four LLMs together) performance of Method 1. 
Prompt 1.1: improving the cognitive dimension shows the best improvement. Prompt 2: improving all dimensions and Prompt 4: improving the lacking dimension perform equally well. Prompt 3: improving on three dimensions sequentially under-performs other strategies, but still improves over the Naive Prompt. The SFT then RLHF DPO fine-tuning shows a consistent performance improvement over SFT only. 

\begin{table*}
    \centering
    \scriptsize
    \renewcommand{\arraystretch}{1.2}  
    \setlength{\tabcolsep}{4.95pt}
    \begin{tabular}{l|l|c|c|c|c|c|c|c|c|c|c|c|c|c|c}
        \toprule
        \hline
         \multicolumn{2}{l|}{\bf Distillation Method}         & \multicolumn{7}{c|}{Improvement over Human Responses} & \multicolumn{7}{c}{ Improvement over LLM Initial Responses}  \\ \hline 
         \multicolumn{2}{l|}{\bf Prompting Strategy}          & N & 1.1 & 1.2 & 1.3 & 2 & 3 & 4 & N & 1.1 & 1.2 & 1.3 & 2 & 3 & 4 \\ \hline 
         GPT-4o teaches LLaMA-3.1-8B,               & SFT & 56.4 & 63.2 & 71.2 & 75.2 & 73.2 & 69.2 & 66.8 & 62.4 & 72.8 & 76.8 & 77.6 & 76.0 & 69.6 & 79.2\\ \cline{2-16} 
         judged by GPT-4o                           & SFT-DPO & 64.8 & 80.4 & 76.8 & 75.6 & 77.6 & 71.2 & 78.8 & 73.6 & 80.8 & 85.2 & 81.6 & 92.8 & 70.8 & 89.6\\ \hline 
         GPT-4o teaches Mistral-7B-v0.3,            & SFT & 67.2 & 87.6 & 88.4 & 75.6 & 90.0 & 72.8 & 91.6 & 89.2 & 94.0 & 92.0 & 94.4 & 98.0 & 92.4 & 93.2\\ \cline{2-16}
         judged by GPT-4o                           & SFT-DPO & 87.6 & 91.6 & 90.4 & 92.4 & 97.2 & 87.2 & 91.2 & 93.2 & 94.4 & 95.2 & 94.0 & 97.6 & 83.6 & 95.2\\ \hline
         GPT-4o teaches LLaMA-3.1-8B,               & SFT & 90.0 & 91.2 & 96.4 & 93.6 & 96.0 & 94.4 & 94.4 & 95.6 & 95.6 & 96.0 & 93.2 & 94.8 & 98.0 & 95.6 \\ \cline{2-16} 
         judged by Gemini                           & SFT-DPO & 98.0 & 96.4 & 97.2 & 98.4 & 97.2 & 95.6 & 98.4 & 96.8 & 97.6 & 98.8 & 96.8 & 99.2 & 96.4 & 99.2\\ \hline 
         GPT-4o teaches Mistral-7B-v0.3,            & SFT & 81.2 & 96.8 & 95.2 & 85.6 & 95.6 & 87.2 & 96.4 & 97.6 & 98.4 & 98.4 & 98.0 & 98.8 & 98.0 & 98.0\\ \cline{2-16}
         judged by Gemini                           & SFT-DPO & 97.2 & 98.8 & 99.2 & 98.8 & 99.2 & 98.8 & 94.8 & 98.4 & 98.0 & 99.6 & 99.6 & 100.0 & 94.4 & 97.6\\ 
        \hline
        \bottomrule
    \end{tabular}
    \caption{Evaluation results of Methods 2 and 3: fine-tuned models' win rate percentages over base models}
    \label{tab:eval}
    \vspace{-0.15in}
\end{table*}

\subsection{Method 3: Targeted Empathy Improvement over LLM Initial Responses}
The third method we develop for empathy distillation from LLMs to SLMs is to utilize the targeted empathy improvement prompts created in Method 2 to bootstrap a distillation dataset without human involvement. In Method 1, the LLM is directly prompted to generate an empathetic response to a dialogue context. Human responses are given in the dataset, but they are not utilized when distilling the LLM. In Method 2, targeted empathy improvement is made over human responses to create the distillation dataset. In Method 3, we investigate the possibility of using LLM generated responses as the initial responses. For each dialogue context, we first query the LLM with the Empathetic Response prompt developed by~\cite{Welivita2024AreLL} to generate the initial response, and then improve over the response with the prompts created in Method 2. 

\begin{figure}[htb]
    \centering
    \includegraphics[width=1\linewidth]{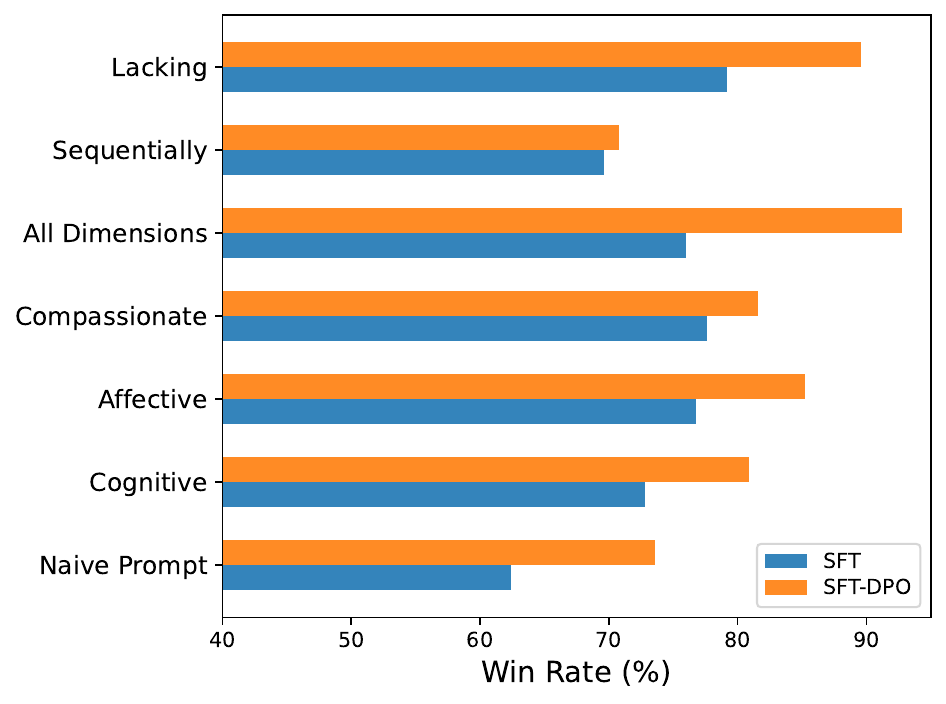}
    \caption{Performance of targeted empathy improvement over LLM generated initial responses}
    \label{fig:llm-initial-response}
    \vspace{-0.2in}
\end{figure}

Since we want to eliminate the reliance on humans in the distillation process, we will not ask humans to give the empathy scores to the initial and improved responses from the LLM. A question that arises then is how to determine what empathetic response examples will make up the SFT and RLHF datasets. Recall that in Methods 1 and 2, the separation of dialogues into the SFT dataset and the RLHF dataset utilize the responses' empathy scores given by humans. However, with the removal of human mediation in the empathy distillation process, we do not have human empathy scores to help partition the SFT and RLHF datasets. Our strategy is to adopt the same exact ratio for the SFT and RLHF datasets in Method 3 as the dataset partition in Methods 2. For the dialogues selected for SFT, we include both the initial and improved responses as separate dialogues in the dataset and for those selected for RLHF, the initial and improved responses form the contrastive pairs.  


Figure~\ref{fig:llm-initial-response} illustrates the performance of targeted empathy improvement over LLM initial responses, with GPT-4o as the LLM and LLaMA-3.1-8B as the SLM. The win rates over the base model are more than 60\% for all prompting strategies with fine-tuning through either SFT only or SFT then RLHF DPO. The two-step fine-tuning achieves the highest win rate of more than 90\% and consistently outperforms SFT-only fine-tuning for all prompting strategies except Prompt 3: Improving the three dimensions of empathy sequentially. Targeted empathy improvement substantially improves over the basic direct prompting by 10+\% in win rate.

\begin{figure}[htb]
    \vspace{-0.15in}
    \centering
    \includegraphics[width=1\linewidth]{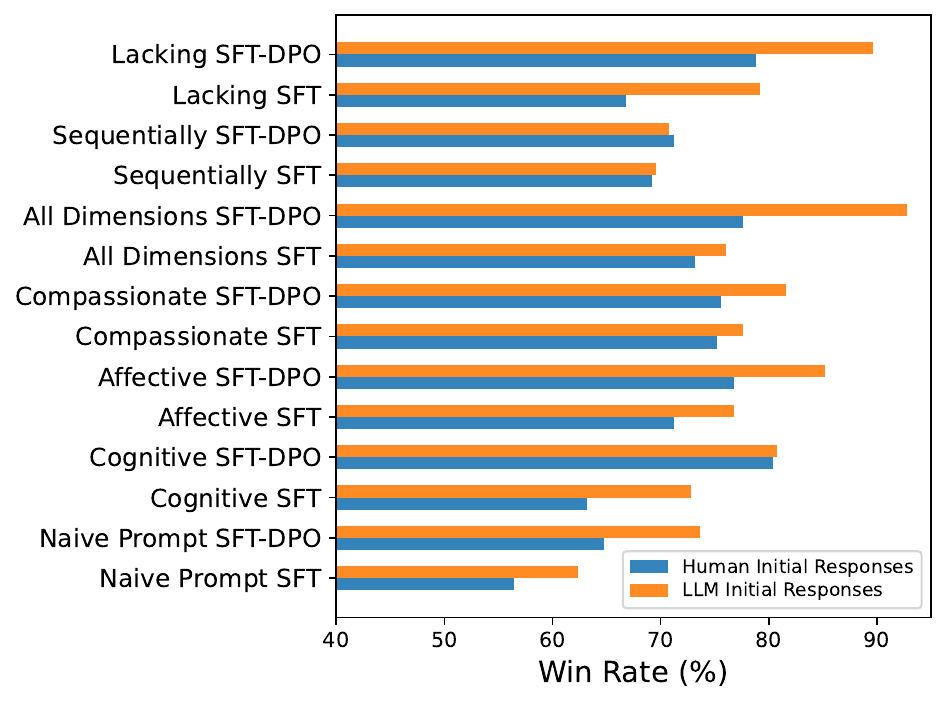}
    \caption{Comparison of targeted empathy improvement over human responses vs. LLM initial responses}
    \label{fig:hir-vs-lir}
    \vspace{-0.2in}
\end{figure}

Figure~\ref{fig:hir-vs-lir} shows the performance comparison of LLaMA-3.1-8B fine-tuned on datasets improved from human responses and from LLM initial responses, respectively, using empathy improvement prompts. It can be observed that the dataset from LLM initial responses consistently outperform that from human responses. There are several possible reasons. (1) As shown by~\cite{Welivita2024AreLL}, state-of-the-art LLMs tend to have better mastery of empathy than humans; therefore, the initial responses from LLMs may already be more empathetic than human responses. (2) The LLM employed for improving the responses may comprehend the LLM initial responses better, possibly making it easier for LLM to improve the responses.

\section{Evaluation and Discussion}


Table~\ref{tab:eval} provides the comprehensive results from our evaluation of Methods 2 and 3: targeted empathy improvement over human responses and LLM initial responses. The leftmost column lists four studies that we conducted: using GPT-4o to teach LLaMA-3.1-8B and Mistral-7B-v0.3 with GPT-4o and Gemini-2.0-Flash as the judges for win-rate evaluation. For each method, we evaluate all five prompting strategies with seven prompts in total.

\begin{figure}[htb]
    \centering
    \vspace{-0.15in}
    \includegraphics[width=1\linewidth]{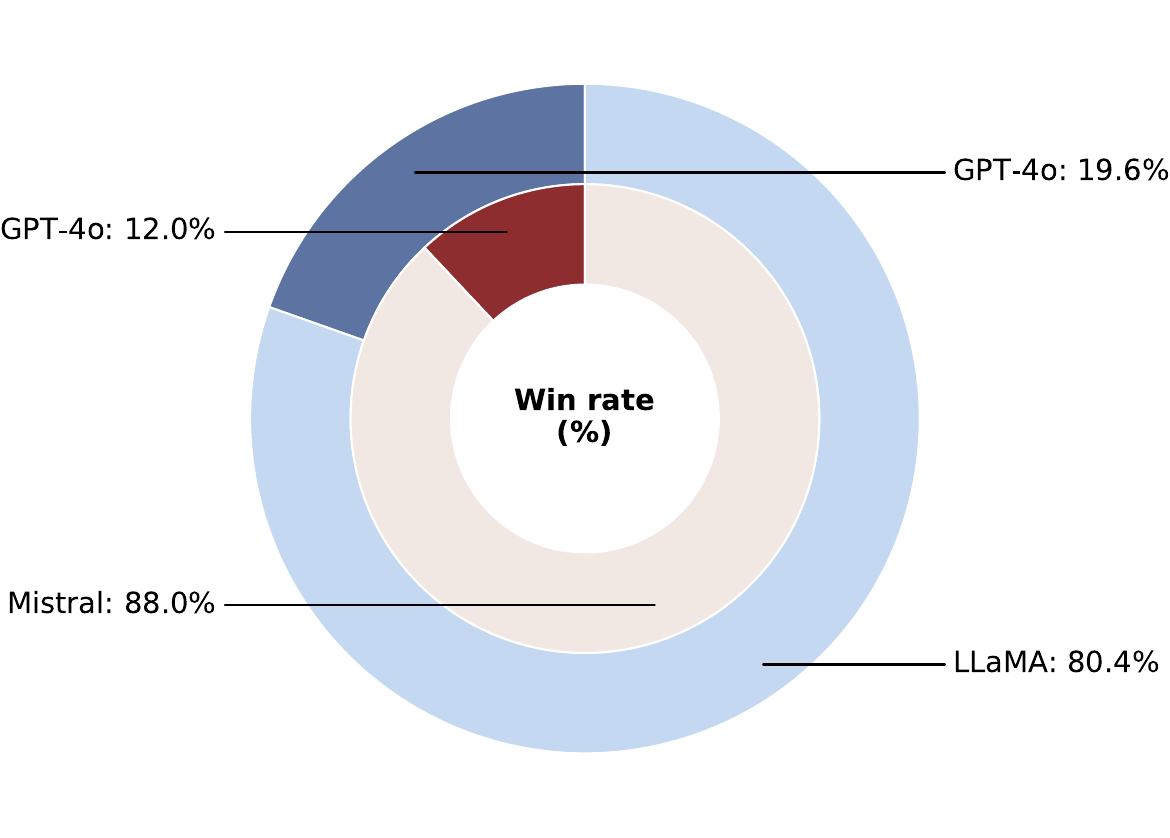}
    \vspace{-0.25in}
    \caption{GPT-4o vs. Fine-tuned LLaMA \& Mistral in empathetic responses as judged by Gemini} 
    \label{fig:win-rate-comp-finetuned}
    \vspace{-0.15in}
\end{figure}

It can be observed that targeted empathy improvement over human or LLM initial responses outperforms Method 1: direct empathy distillation (see Figure~\ref{fig:direction-distillation}) consistently under both SFT only and SFT then RLHF DPO. For different teacher and student combinations, the performances of different prompting strategies vary. This indicates the benefits of our variety of strategies. Though in two combinations the teacher and judge are the same (GPT-4o), these studies show similar trends as the studies where the teacher and judge are different.

Figure~\ref{fig:win-rate-comp-finetuned} illustrates the win rate evaluation pitting GPT-4o against the best fine-tuned LLaMA-3.1-8B and Mistral-7B-v0.3 models. All three models are prompted with the basic direct prompt. Both fine-tuned SLMs outperform GPT-4o in generating empathetic responses, judged by Gemini-2.0-Flash. 

\section{Conclusions}
As knowledge is distilled from LLMs to SLMs, their empathetic capabilities must also be preserved. 
In this paper, we present a comprehensive approach to distilling empathy from LLMs into SLMs. 
First, we developed a two-step fine-tuning process that conducts SFT first with high empathy responses and then applies RLHF DPO with (low, high) empathy response pairs. 
Second, we explored three different methods to distill empathy from LLMs (1) direct empathy distillation, (2) targeted empathy improvement over human responses, and (3) target empathy improvement over LLM initial responses. 
Third, for targeted empathy improvement, we explored four prompting strategies, all of which demonstrate significant improvement in distilling empathy over direct prompting and achieve varying success when combined with different initial responses and SLMs. 
Our study shows that distilling empathy from LLMs into SLMs is not only feasible but can also be done extremely effectively.

\clearpage
\section*{Limitations}
So far, we have only conducted the win rate evaluation on the empathy of our distilled SLMs with LLMs as the judges, meaning that our evaluations are subject to LLMs' biases and hallucinations. To combat this limitation and given that such SLMs interact frequently with humans, human evaluation is an essential step towards practical application of our presented approach. As an immediate follow-up, we will conduct human studies on rating the empathy of the responses generated by the distilled SLMs. We plan to utilize the human empathy rating scheme developed by~\cite{Welivita2024AreLL} to evaluate the effectiveness of empathy distillation.    

\section*{Ethical Statement}
The aim of our research is to improve the empathy of SLMs by distilling empathy from state-of-the-art LLMs. As in any distillation, the shortcomings of LLMs can often propagate into the distilled SLMs. The distillation methods proposed, while designed for distilling empathy, can be abused as methods for distilling negative behaviors of certain LLMs into SLMs, which we strongly advocate against.

\section*{Acknowledgments}
Henry J. Xie is partially supported by a gift from Youth for Empathetic AI. Dr. Kunpeng Liu is supported by the National Science Foundation (NSF) via the grant numbers 2426339 and 2348485.

\bibliography{anthology,custom}

\begin{thebibliography}{24}
\providecommand{\natexlab}[1]{#1}

\bibitem[{Cui et~al.(2023)Cui, Yang, Liang, Wu, and Lin}]{cui2023rise}
Zhe Cui, Fan Yang, Shuo Liang, Yunhe Wu, and Dahua Lin. 2023.
\newblock \href {https://arxiv.org/abs/2306.07701} {The rise of on-device {AI}: A survey of challenges and techniques}.
\newblock \emph{arXiv preprint, abs/2306.07701}.

\bibitem[{Davis(1983)}]{davis1983}
Mark~H. Davis. 1983.
\newblock \href {https://doi.org/10.1037/0022-3514.44.1.113} {Measuring individual differences in empathy: Evidence for a multidimensional approach}.
\newblock \emph{Journal of Personality and Social Psychology}, 44(1):113--126.

\bibitem[{DeepSeek-AI(2025)}]{DeepSeekAI2025DeepSeekR1IR}
DeepSeek-AI. 2025.
\newblock \href {https://api.semanticscholar.org/CorpusID:275789950} {{DeepSeek-R1}: Incentivizing reasoning capability in {LLMs} via reinforcement learning}.
\newblock \emph{Arxiv preprint}, abs/2501.12948.

\bibitem[{{Google}(2023)}]{google2023gemini}
{Google}. 2023.
\newblock \href {https://blog.google/technology/ai/google-gemini-ai/} {Introducing {Gemini}: Our largest and most capable {AI} model}.
\newblock Goolge Technology Blog.

\bibitem[{{Google}(2024)}]{google2024gemini2}
{Google}. 2024.
\newblock \href {https://blog.google/technology/google-deepmind/google-gemini-ai-update-december-2024/} {Introducing {Gemini} 2.0: our new ai model for the agentic era}.
\newblock Goolge Technology Blog.

\bibitem[{Ma et~al.(2020)Ma, Nguyen, Xing, and Cambria}]{Ma2020}
Yukun Ma, Khanh~Linh Nguyen, Frank~Z. Xing, and Erik Cambria. 2020.
\newblock \href {https://doi.org/10.1016/j.inffus.2020.06.011} {A survey on empathetic dialogue systems}.
\newblock \emph{Information Fusion}, 64:50--70.

\bibitem[{{Meta AI}(2024)}]{meta2024LLaMA3}
{Meta AI}. 2024.
\newblock \href {https://ai.meta.com/blog/meta-llama-3-1/} {Introducing {LLaMA} 3.1: Our most capable models to date}.
\newblock Meta AI Blog.

\bibitem[{{Mistral AI}(2023)}]{mistral2023mistral7b}
{Mistral AI}. 2023.
\newblock \href {https://mistral.ai/news/announcing-mistral-7b} {Mistral 7b}.
\newblock Mistral AI News.

\bibitem[{{Mistral AI}(2024)}]{mistralai2024mixtral}
{Mistral AI}. 2024.
\newblock \href {https://arxiv.org/abs/2401.04088} {Mixtral of experts}.
\newblock \emph{Arxiv preprint}, abs/2401.04088.

\bibitem[{{OpenAI}(2023)}]{openai2023gpt4}
{OpenAI}. 2023.
\newblock \href {https://arxiv.org/abs/2303.08774} {{GPT-4} technical report}.
\newblock \emph{Arxiv preprint}, abs/2303.08774.

\bibitem[{OpenAI(2024)}]{openai2024gpt4o}
OpenAI. 2024.
\newblock \href {https://openai.com/index/gpt-4o} {{GPT-4o}: Openai’s new flagship model}.
\newblock OpenAI Blog.

\bibitem[{{Qualcomm}(2023)}]{qualcomm2023ondevice}
{Qualcomm}. 2023.
\newblock \href {https://www.qualcomm.com/news/onq/2023/09/ai-on-the-edge-the-latest-on-device-ai-insights-and-trends} {On-device {AI}: Trends, challenges, and opportunities}.
\newblock Qualcomm OnQ Blog.

\bibitem[{Raamkumar and Yang(2022)}]{Raamkumar2022}
Aravind~Sesagiri Raamkumar and Yinping Yang. 2022.
\newblock \href {https://doi.org/10.1109/TAFFC.2022.3226693} {Empathetic conversational systems: A review of current advances, gaps, and opportunities}.
\newblock \emph{IEEE Transactions on Affective Computing}.

\bibitem[{Rafailov et~al.(2023)Rafailov, Sharma, Mitchell, Manning, Ermon, and Finn}]{rafailov2023dpo}
Rafael Rafailov, Archit Sharma, Eric Mitchell, Christopher~D. Manning, Stefano Ermon, and Chelsea Finn. 2023.
\newblock \href {https://arxiv.org/abs/2305.18290} {Direct preference optimization: Your language model is secretly a reward model}.
\newblock In \emph{Advances in Neural Information Processing Systems (NeurIPS)}.

\bibitem[{Rashkin et~al.(2018)Rashkin, Smith, Li, and Boureau}]{Rashkin2018TowardsEO}
Hannah Rashkin, Eric~Michael Smith, Margaret Li, and Y-Lan Boureau. 2018.
\newblock \href {https://api.semanticscholar.org/CorpusID:195069365} {Towards empathetic open-domain conversation models: A new benchmark and dataset}.
\newblock In \emph{Annual Meeting of the Association for Computational Linguistics (ACL)}.

\bibitem[{Sreenivas et~al.(2024)Sreenivas, Muralidharan, Joshi, Chochowski, Patwary, Shoeybi, Catanzaro, Kautz, and Molchanov}]{Sreenivas2024LLMPA}
Sharath~Turuvekere Sreenivas, Saurav Muralidharan, Raviraj Joshi, Marcin Chochowski, Mostofa Patwary, Mohammad Shoeybi, Bryan Catanzaro, Jan Kautz, and Pavlo Molchanov. 2024.
\newblock \href {https://api.semanticscholar.org/CorpusID:271915771} {{LLM} pruning and distillation in practice: The {Minitron} approach}.
\newblock \emph{Arxiv preprint}, abs/2408.11796.

\bibitem[{Touvron et~al.(2023)Touvron, Lavril, Izacard, Martinet, Lachaux, Lacroix, Rozière, Goyal, Hambro, Azhar, Rodriguez, Joulin, and Lample}]{touvron2023LLaMA}
Hugo Touvron, Thibaut Lavril, Gautier Izacard, Xavier Martinet, Marie-Anne Lachaux, Timothée Lacroix, Baptiste Rozière, Naman Goyal, Eric Hambro, Faisal Azhar, Edouard~Grave Rodriguez, Armand Joulin, and Guillaume Lample. 2023.
\newblock \href {https://arxiv.org/abs/2307.09288} {{LLaMA} 2: Open foundation and fine-tuned chat models}.
\newblock \emph{Arxiv preprint}, abs/2307.09288.

\bibitem[{Wang et~al.(2025)Wang, Tang, Guo, Meng, Wang, Wang, and Jia}]{Wang_2025}
Xubin Wang, Zhiqing Tang, Jianxiong Guo, Tianhui Meng, Chenhao Wang, Tian Wang, and Weijia Jia. 2025.
\newblock \href {https://doi.org/10.1145/3724420} {Empowering edge intelligence: A comprehensive survey on on-device {AI} models}.
\newblock \emph{ACM Computing Surveys}, 57(9):1–39.

\bibitem[{Welivita and Pu(2020)}]{welivita-pu-2020-taxonomy}
Anuradha Welivita and Pearl Pu. 2020.
\newblock \href {https://doi.org/10.18653/v1/2020.coling-main.429} {A taxonomy of empathetic response intents in human social conversations}.
\newblock In \emph{Proceedings of the 28th International Conference on Computational Linguistics}, pages 4886--4899, Barcelona, Spain (Online). International Committee on Computational Linguistics.

\bibitem[{Welivita and Pu(2024)}]{Welivita2024AreLL}
Anuradha Welivita and Pearl Pu. 2024.
\newblock \href {https://api.semanticscholar.org/CorpusID:270357813} {Are large language models more empathetic than humans?}
\newblock \emph{Arxiv preprint}, abs/2406.05063.

\bibitem[{Xie et~al.(2024)Xie, Zhang, Zhang, and Liu}]{Xie2024ScoringWL}
Henry~J. Xie, Jinghan Zhang, Xinhao Zhang, and Kunpeng Liu. 2024.
\newblock \href {https://api.semanticscholar.org/CorpusID:275133806} {Scoring with large language models: A study on measuring empathy of responses in dialogues}.
\newblock \emph{2024 IEEE International Conference on Big Data (BigData)}, pages 7433--7437.

\bibitem[{Xie and Agrawal(2023)}]{DBLP:conf/wassa/XieA23}
Justin~J. Xie and Ameeta Agrawal. 2023.
\newblock \href {https://doi.org/10.18653/V1/2023.WASSA-1.7} {Emotion and sentiment guided paraphrasing}.
\newblock In \emph{Proceedings of the 13th Workshop on Computational Approaches to Subjectivity, Sentiment, {\&} Social Media Analysis (WASSA@ACL)}, pages 58--70. Association for Computational Linguistics (ACL).

\bibitem[{Xu et~al.(2024)Xu, Li, Tao, Shen, Cheng, Li, Xu, Tao, and Zhou}]{Xu2024ASO}
Xiaohan Xu, Ming Li, Chongyang Tao, Tao Shen, Reynold Cheng, Jinyang Li, Can Xu, Dacheng Tao, and Tianyi Zhou. 2024.
\newblock \href {https://api.semanticscholar.org/CorpusID:267760021} {A survey on knowledge distillation of large language models}.
\newblock \emph{Arxiv preprint}, abs/2402.13116.

\bibitem[{Zheng et~al.(2024)Zheng, Zhang, Zhang, Ye, Luo, and Ma}]{Zheng2024LLaMAFactoryUE}
Yaowei Zheng, Richong Zhang, Junhao Zhang, Yanhan Ye, Zheyan Luo, and Yongqiang Ma. 2024.
\newblock \href {https://api.semanticscholar.org/CorpusID:268536974} {{LLaMAFactory}: Unified efficient fine-tuning of 100+ language models}.
\newblock \emph{Arxiv preprint}, abs/2403.13372.

\end{thebibliography}

\clearpage
\appendix

\section{Appendix}
\label{sec:appendix}

Below please find the full prompts that we developed in support of Distillation Methods 2 and 3.  
\begin{center}
    \fbox{\rule[-.15cm]{0cm}{0cm}
    \begin{minipage}[t]{2.9in}
       {\small
        \noindent{\bf Naive Prompt}
        \\
        Below is a response to a given speaker utterance in a given context. Generate a new improved empathetic response, using on average 28 words and a maximum of 97 words, that is of higher empathetic quality and also retains the original meaning, intention, and emotion of the original response.
        }
    \end{minipage}
}
\end{center}

\begin{center}
    \fbox{\rule[-.15cm]{0cm}{0cm}
    \begin{minipage}[t]{2.9in}
        {\small
        \noindent{\bf Prompt 1.1: Improve along the Cognitive Dimension}\\[1ex]
        Below is a response to a given speaker utterance in a given context. Generate a new improved empathetic response, using on average 28 words and a maximum of 97 words, that is of higher empathetic quality and also retains the original meaning, intention, and emotion of the original response.
        \\
        Your higher quality response should be improved specifically along the cognitive dimension of empathy. 
        \\
        Cognitive empathy is the ability to understand another person's thoughts, beliefs, and intentions. It is being able to see the world through their eyes and understand their point of view.
        \\[1ex]
        \noindent{\bf Prompt 1.2: Improve along the Affective Dimension}\\[1ex]
        Below is a response to a given speaker utterance in a given context. Generate a new improved empathetic response, using on average 28 words and a maximum of 97 words, that is of higher empathetic quality and also retains the original meaning, intention, and emotion of the original response.
        \\
        Your higher quality response should be improved specifically along the affective (emotional) dimension of empathy. 
        \\
        Affective empathy is the ability to experience the emotions of another person. It is feeling what they are feeling, both positive and negative.
        \\[1ex]
        \noindent{\bf Prompt 1.3: Improve along the Compassionate Dimension}\\[1ex]
        Your higher quality response should be improved specifically along the compassionate dimension of empathy. 
        \\
        Compassionate empathy is the ability to not only understand and share another person's feelings, but also to be moved to help if needed. It involves a deeper level of emotional engagement than cognitive empathy, prompting action to alleviate another's distress or suffering.
        }
    \end{minipage}
}
\end{center}

\begin{center}
    \fbox{\rule[-.15cm]{0cm}{0cm}
    \begin{minipage}[t]{2.9in}
       {\small
        \noindent{\bf Prompt 2: Improve All Three Dimensions of Empathy}
        \\
        Below is a response to a given speaker utterance in a given context. Generate a new improved empathetic response, using on average 28 words and a maximum of 97 words, that is of higher empathetic quality and also retains the original meaning, intention, and emotion of the original response. 
        \\
        Your higher quality response should be improved along the three dimensions of empathy: cognitive, affective(emotional), and compassionate empathy. 
        \\
        Cognitive empathy is the ability to understand another person's thoughts, beliefs, and intentions. It is being able to see the world through their eyes and understand their point of view. Affective empathy is the ability to experience the emotions of another person. It is feeling what they are feeling, both positive and negative. Compassionate empathy is the ability to not only understand and share another person's feelings, but also to be moved to help if needed. It involves a deeper level of emotional engagement than cognitive empathy, prompting action to alleviate another's distress or suffering.
        }
    \end{minipage}
}
\end{center}

\begin{center}
    \fbox{\rule[-.15cm]{0cm}{0cm}
    \begin{minipage}[t]{2.9in}
       {\small
        \noindent{\bf Prompt 3: Improve Three Dimensions Sequentially}\\[1ex]
        Below is a response to a given speaker utterance in a given context. Generate a new improved empathetic response, using on average 28 words and a maximum of 97 words, that is of higher empathetic quality and also retains the original meaning, intention, and emotion of the original response. Your higher quality response should be improved specifically along the cognitive dimension of empathy. Cognitive empathy is the ability to understand another person's thoughts, beliefs, and intentions. It is being able to see the world through their eyes and understand their point of view.
        \\[1ex]
        Generate a new improved empathetic response, using on average 28 words and a maximum of 97 words, that is of higher empathetic quality and also retains the original meaning, intention, and emotion of the original response. Your higher quality response should be improved specifically along the affective(emotional) dimension of empathy. Affective empathy is the ability to experience the emotions of another person. It is feeling what they are feeling, both positive and negative.
        \\[1ex]
        Generate a new improved empathetic response, using on average 28 words and a maximum of 97 words, that is of higher empathetic quality and also retains the original meaning, intention, and emotion of the original response. Your higher quality response should be improved specifically along the compassionate dimension of empathy. Compassionate empathy is the ability to not only understand and share another person's feelings, but also to be moved to help if needed. It involves a deeper level of emotional engagement than cognitive empathy, prompting action to alleviate another's distress or suffering.
        }
    \end{minipage}
}
\end{center}

\begin{center}
    \fbox{\rule[-.15cm]{0cm}{0cm}
    \begin{minipage}[t]{2.9in}
       {\small
        \noindent{\bf Prompt 4: Identify the Lacking Dimension}
        \\
        Below is a response to a given speaker utterance in a given context. Generate a new improved empathetic response, using on average 28 words and a maximum of 97 words, that is of higher empathetic quality and also retains the original meaning, intention, and emotion of the original response. 
        \\ 
        \\
        In the process of generating a higher quality empathetic response, you should identify the dimension of empathy(cognitive, affective, and compassionate dimensions) that the original response lacks most of, and specifically improve along the lines of the dimension you identified. Cognitive empathy is the ability to understand another person's thoughts, beliefs, and intentions. It is being able to see the world through their eyes and understand their point of view. Affective empathy is the ability to experience the emotions of another person. It is feeling what they are feeling, both positive and negative. Compassionate empathy is the ability to not only understand and share another person's feelings, but also to be moved to help if needed. It involves a deeper level of emotional engagement than cognitive empathy, prompting action to alleviate another's distress or suffering.
        }
    \end{minipage}
}
\end{center}

\end{document}